\documentclass[11pt,twocolumn]{article}

\usepackage[margin=1in]{geometry}
\usepackage{graphicx}
\usepackage[numbers]{natbib}
\usepackage{listings}
\usepackage{booktabs}
\usepackage[colorlinks=true,linkcolor=blue!60!black,citecolor=blue!60!black,urlcolor=blue!60!black]{hyperref}
\hypersetup{
  pdftitle={Generative Ontology: When Structured Knowledge Learns to Create},
  pdfauthor={Benny Cheung},
  pdfsubject={Ontology-constrained structured generation for creative AI systems},
  pdfkeywords={generative ontology, structured generation, ontology engineering, LLM, multi-agent systems, game design, DSPy, process philosophy},
  pdfcreator={LaTeX with hyperref}
}
\usepackage{xurl}
\usepackage{xcolor}
\usepackage{amsmath}
\usepackage{caption}
\usepackage{enumitem}
\usepackage{microtype}
\usepackage{tabularx}
\usepackage{titlesec}
\usepackage{multirow}
\titleformat{\section}{\Large\bfseries\raggedright}{\thesection}{1em}{}
\titleformat{\subsection}{\large\bfseries\raggedright}{\thesubsection}{1em}{}

\definecolor{codebg}{HTML}{F5F5F5}
\definecolor{codeframe}{HTML}{CCCCCC}
\definecolor{keyword}{HTML}{0000AA}
\definecolor{string}{HTML}{008800}
\definecolor{comment}{HTML}{888888}

\newcommand{\etal}{\emph{et al.}}

\title{Generative Ontology:\\When Structured Knowledge Learns to Create}

\author{
  Benny Cheung\\
  Dynamind Research\\
  \texttt{btscheung@dynamindresearch.com}
}

\date{}

\begin{document}
\maketitle

% ============================================================
% ABSTRACT
% ============================================================
\begin{abstract}
Traditional ontologies excel at describing domain structure but cannot
generate novel artifacts.  Large language models generate fluently but
produce outputs that lack structural validity, hallucinating mechanisms
without components, goals without end conditions.  We introduce
\emph{Generative Ontology}, a framework that synthesizes these
complementary strengths: ontology provides the grammar; the LLM provides
the creativity.

Generative Ontology encodes domain knowledge as executable Pydantic
schemas that constrain LLM generation via DSPy signatures.  A
multi-agent pipeline assigns specialized roles to different ontology
domains: a Mechanics Architect designs game systems, a Theme Weaver
integrates narrative, a Balance Critic identifies exploits, each
carrying a professional ``anxiety'' that prevents shallow, agreeable
outputs.  Retrieval-augmented generation grounds novel designs in
precedents from existing exemplars, while iterative validation ensures
coherence between mechanisms and components.

We demonstrate the framework through \textsc{GameGrammar}, a system for
generating complete tabletop game designs, and present three empirical
studies.  An \emph{ablation study} (120 designs, 4 conditions) shows
that multi-agent specialization produces the largest creative quality
gains (fun $d = 1.12$, strategic depth $d = 1.59$; both $p < .001$),
while schema validation eliminates structural errors ($d = 4.78$).  A
\emph{benchmark comparison} against 20 published board games reveals
structural parity but a bounded creative gap (fun $d = 1.86$): generated
designs score 7--8 while published games score 8--9, with tension/drama
and social interaction already at parity.  A \emph{test-retest
reliability} study (ICC analysis, 50 evaluations) validates the
LLM-based evaluator, with 7 of 9 metrics achieving Good-to-Excellent
reliability (ICC 0.836--0.989).

The pattern generalizes beyond games.  Any domain with expert vocabulary,
validity constraints, and accumulated exemplars is a candidate for
Generative Ontology.  Constraints do not limit creativity but enable it:
just as grammar makes poetry possible, ontology makes structured
generation possible.
\end{abstract}

% ============================================================
% 1. INTRODUCTION
% ============================================================
\section{Introduction}
\label{sec:introduction}

Ontologies describe what \emph{is}.  They define concepts, relationships,
and constraints that codify expert understanding of a domain.  In our
earlier work on tabletop game ontology~\cite{cheung2025tabletop}, we built
a structured vocabulary that could decompose \emph{Catan} into
mechanisms (resource trading, modular board, dice-driven production),
components (hex tiles, resource cards, settlements), and player dynamics
(competitive, negotiation-heavy, variable player count).  That ontology
gave us a precise language for analysis.

But analysis is not creation.

Understanding the structure of sonnets does not make one a poet.  Knowing
the rules of chess does not generate new games.  A traditional ontology,
for all its precision, remains passive, a tool for describing what
already exists, not for imagining what could be.  We ask: what happens
when we transform that passive vocabulary into an active grammar?

\subsection{The Problem: Structural Hallucination}

Large language models (LLMs) excel at fluent generation.  Ask an LLM to
``design a deck-building game set in a haunted mansion'' and it will
produce paragraphs of creative output: themes, mechanisms, components,
and victory conditions flowing freely.  Yet this fluency hides a deeper
problem: \textbf{LLMs generate without understanding structure}.

Consider the output for such a prompt: ``\emph{Players explore
Ravenshollow Manor, collecting ghost cards and building their decks.  The
game features a unique `fear mechanic' where players manage their sanity.
Victory goes to whoever escapes with the most treasure}\ldots''  This
sounds plausible.  But what cards exist in the starting deck?  How do
players acquire new cards?  What triggers the end of the game?  How does
the ``fear mechanic'' interact with deck composition?  The LLM has
generated the \emph{appearance} of a game design without the
\emph{substance}.  Handed to a playtester, it would collapse into
confusion.  This is the hallucination problem applied to creative
domains: just as LLMs hallucinate facts, they hallucinate
structures, producing outputs that \emph{sound} valid without
\emph{being} valid.

\begin{figure*}[t]
\centering
\includegraphics[width=0.85\textwidth]{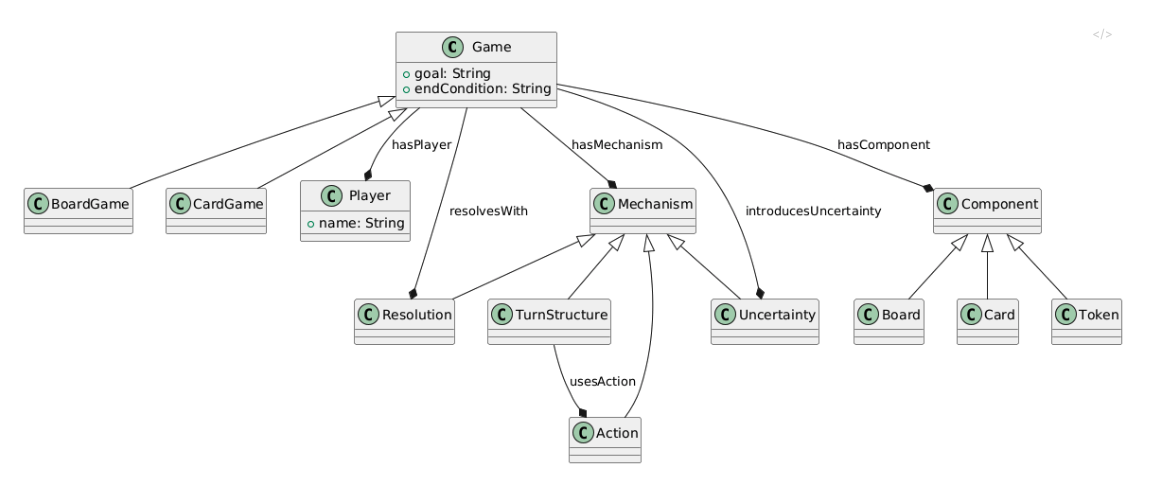}
\caption{A sample Game Ontology expressed as an OntoUML class diagram.
The \texttt{Game} root entity decomposes into \texttt{Player},
\texttt{Mechanism}, and \texttt{Component} aggregates, each with
typed relationships (\emph{hasPlayer}, \emph{hasMechanism},
\emph{hasComponent}) that mirror the domain structure a Generative
Ontology must capture and enforce.}
\label{fig:hero}
\end{figure*}

\subsection{Complementary Limitations}

Table~\ref{tab:comparison} summarizes the impasse.  Traditional ontology
has structure without creativity; the LLM has creativity without
structure.  Neither alone solves the problem of generating novel, valid
designs.

\begin{table}[t]
\centering
\caption{Traditional ontology versus pure LLM generation.}
\label{tab:comparison}
\small
\begin{tabularx}{\columnwidth}{lXX}
\toprule
& \textbf{Strength} & \textbf{Weakness} \\
\midrule
\textbf{Ontology} & Precise, structured, validated & Cannot generate novel outputs \\
\textbf{LLM}      & Creative, fluent, abundant     & Unstructured, hallucinated \\
\bottomrule
\end{tabularx}
\end{table}

The insight of Generative Ontology is that these limitations are
\emph{complementary}.  What ontology lacks, LLMs provide.  What LLMs
lack, ontology provides.  The synthesis, structured generation guided by
ontological schemas, produces outputs that are both creative and valid.

\subsection{Paper Overview}

We introduce \emph{Generative Ontology} and develop it through the
domain of tabletop game design.  Section~\ref{sec:related} reviews
related work.  Section~\ref{sec:synthesis} defines the framework and its
philosophical grounding.  Sections~\ref{sec:schema}--\ref{sec:validation}
develop the technical architecture: executable schemas, DSPy
operationalization, multi-agent pipelines, retrieval augmentation, and
validation contracts.  Section~\ref{sec:evaluation} presents three
empirical studies evaluating the framework.
Section~\ref{sec:casestudy} illustrates the pipeline with a complete
case study.  Section~\ref{sec:generalizability} argues the pattern
generalizes beyond games.  Section~\ref{sec:conclusion} concludes with
future directions.

% ============================================================
% 2. RELATED WORK
% ============================================================
\section{Related Work}
\label{sec:related}

\subsection{Ontologies and LLMs}

Ontology development has a well-established
methodology~\cite{noy2001ontology}, but recent research at the
intersection of ontologies and LLMs proceeds primarily in the
\emph{opposite} direction from our approach.  DRAGON-AI uses
retrieval-augmented LLM generation to construct biomedical ontologies
from text corpora~\cite{toro2024dragon}, and OLLM demonstrates
end-to-end ontology learning with language models.  These systems use
LLMs \emph{to generate ontologies}; we use ontologies \emph{to guide
generation}.

Mehenni and Zouaq~\cite{mehenni2024ontology} demonstrate
ontology-constrained generation for clinical summaries, the closest
methodological precedent to our work, though their goal is
hallucination reduction in summarization rather than structured
creativity.  Cooper~\cite{cooper2024constrained} provides foundational
work on constrained decoding that underpins our structured output
approach.

\subsection{LLMs in Game Design}

Gallotta~\etal~\cite{gallotta2024llm} survey LLM applications in games
broadly, while Becker~\etal~\cite{becker2025boardwalk} present
Boardwalk, a framework for generating board game \emph{code} from
existing rules, generating implementations rather than designs.
Maleki and Zhao~\cite{maleki2024pcg} survey procedural content
generation methods, noting emerging LLM integration but without
ontological constraints.

\subsection{LLM-as-Judge Evaluation}

Recent work has established LLMs as evaluators of creative
output~\cite{zheng2024judging}.  Concerns about evaluator reliability
motivate our test-retest analysis: Shankar~\etal~\cite{shankar2024validates}
argue that ``who validates the validators'' is a critical question for
LLM-based evaluation pipelines.  We address this by reporting
intraclass correlation coefficients (ICC) for all evaluation
metrics~\cite{koo2016guideline,shrout1979intraclass}, demonstrating that
our evaluator is a reliable instrument rather than assuming so.

\subsection{Gap}

Our contribution fills a specific gap: using ontology as a generative
grammar for structured creative output, operationalized through
executable schemas, multi-agent pipelines, and retrieval augmentation,
with rigorous empirical evaluation including ablation, benchmarking, and
evaluator validation.

% ============================================================
% 3. GENERATIVE ONTOLOGY: THE SYNTHESIS
% ============================================================
\section{Generative Ontology: The Synthesis}
\label{sec:synthesis}

If traditional ontology is a map, Generative Ontology is a map that
knows how to build new cities.

\begin{quote}
\textbf{Generative Ontology} is the practice of encoding domain
knowledge as executable schemas that constrain and guide AI generation,
transforming static knowledge representation into a creative engine.
\end{quote}

When we built our tabletop game ontology~\cite{cheung2025tabletop}, we
created a structured vocabulary: game types, mechanisms, components,
player interactions.  This vocabulary enabled \emph{analysis} of games
such as \emph{Catan} and \emph{Dune: Imperium} with precision.  The
insight of Generative Ontology is that the same structure enabling
understanding can enable creation.  The categories that constrain
analysis become the scaffolding for generation.

\subsection{The Grammar of Games}

A poet does not experience grammar as a limitation.  Grammar is not what
prevents poetry; it is what makes poetry \emph{possible}.  Without the
structure of syntax, semantics, and form, there would be no sonnets, no
haiku, no free verse pushing against convention.  The rules are not the
enemy of creativity; they are its condition of possibility.

The same principle applies to game design.  When we ask an unconstrained
LLM to ``design a board game about space exploration,'' we receive
verbose descriptions that sound plausible but collapse under scrutiny.
The AI might describe a ``resource management mechanic'' without
specifying what resources exist, how they flow, or what makes acquiring
them interesting.  Generative Ontology provides the grammar.  When we
encode the game ontology as a schema, we give the LLM the structural
vocabulary to be creative \emph{coherently}.  The schema declares: every
game must have a goal, an end condition, mechanisms that create player
choices, components that instantiate those mechanisms.  Within those
constraints, infinitely many games are possible.  Without them, no valid
game emerges.

\begin{figure*}[t]
\centering
\includegraphics[width=0.95\textwidth]{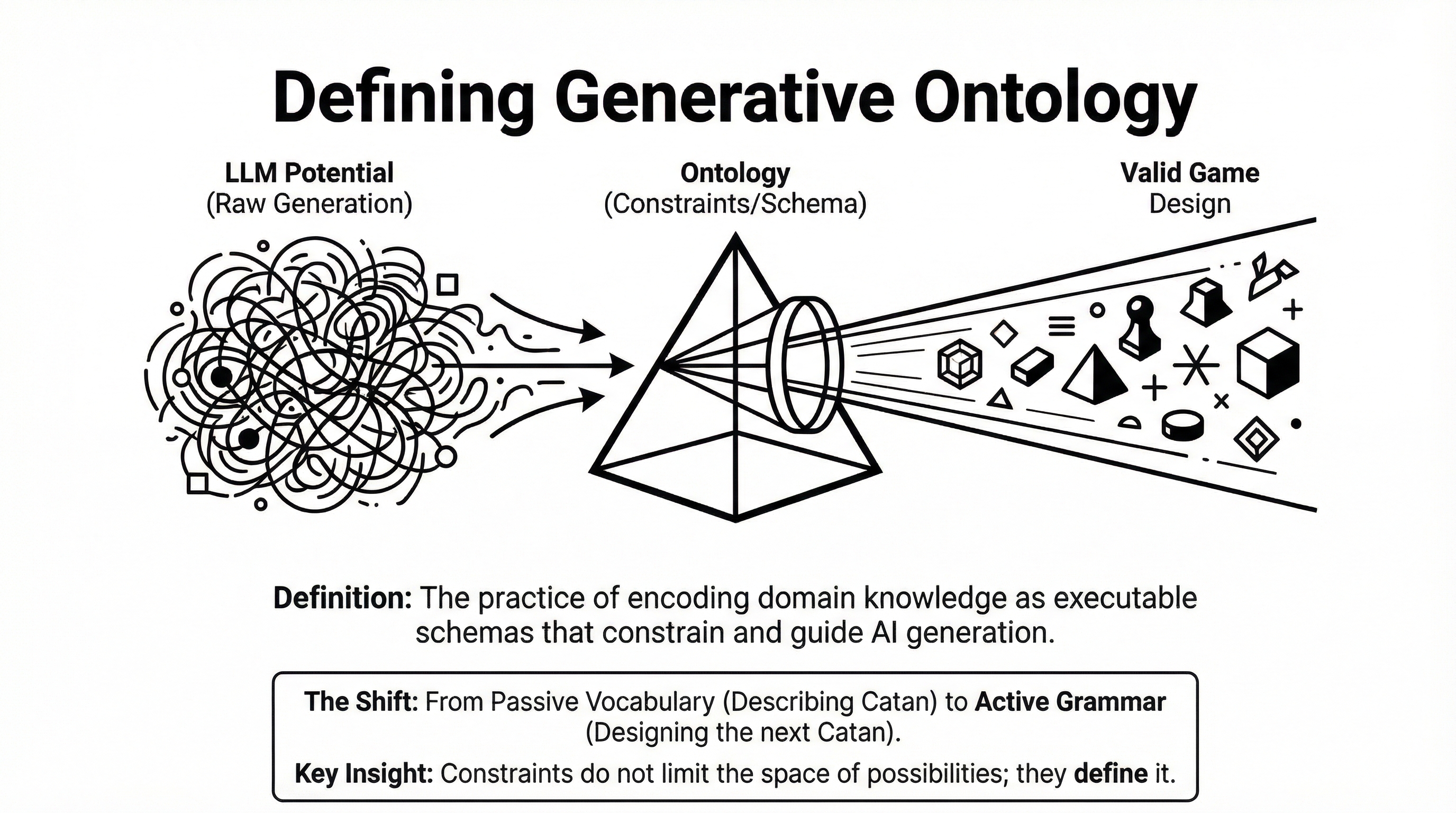}
\caption{LLM potential plus ontology constraints yields valid design.
Generative Ontology transforms a passive vocabulary for description into
an active grammar for creation.}
\label{fig:definition}
\end{figure*}

Table~\ref{tab:with-without} illustrates the difference concretely.

\begin{table}[t]
\centering
\caption{Generation without versus with Generative Ontology.}
\label{tab:with-without}
\small
\begin{tabular}{@{}p{0.45\columnwidth}p{0.45\columnwidth}@{}}
\toprule
\textbf{Without Ontology} & \textbf{With Generative Ontology} \\
\midrule
\raggedright ``Players collect resources and build things'' &
\raggedright Goal: first to 10~VP via settlements \tabularnewline
\raggedright ``There's some kind of trading'' &
\raggedright Mechanism: resource trading with 4:1 bank or negotiated rates \tabularnewline
\raggedright ``The game ends eventually'' &
\raggedright End Condition: when any player reaches 10~VP or development deck exhausts \tabularnewline
Fluent but vague & Structured and playable \\
\bottomrule
\end{tabular}
\end{table}

\subsection{Philosophical Grounding}

For readers familiar with our earlier exploration of process philosophy
for AI agent design~\cite{cheung2026process}, Generative Ontology has a
natural grounding in Alfred North Whitehead's
metaphysics~\cite{whitehead1929process}.  Whitehead distinguished
between \emph{eternal objects} (pure forms, patterns, and structures
that exist as potentials) and \emph{actual occasions} (the concrete
events where eternal objects find expression).  Our game ontology is a
collection of eternal objects: the abstract patterns of worker placement,
deck building, area control.  These patterns have no reality until they
are instantiated in actual games.  Generative Ontology is the process of
moving from eternal objects to actual occasions: the ontology provides
the forms; generation produces the instances.  Creativity, in
Whitehead's view, is precisely this novel synthesis of eternal objects
into actual occasions that have never existed
before~\cite{barker2024artificial}.

% ============================================================
% 4. FROM ONTOLOGY TO EXECUTABLE SCHEMA
% ============================================================
\section{From Ontology to Executable Schema}
\label{sec:schema}

Philosophy illuminates the path; engineering builds the road.  The key
insight is surprisingly direct: ontology classes map naturally to schema
definitions, and schema definitions become the structured outputs that
constrain LLM generation.

\subsection{The Transformation}

Our tabletop game ontology~\cite{cheung2025tabletop,engelstein2020building}
defined four core concepts: \emph{Game} (the container with goal and end
condition), \emph{Mechanism} (turn structure, actions, resolution,
uncertainty), \emph{Component} (board, cards, tokens), and
\emph{Player} (roles and interactions).  Each of these ontology classes
becomes a Pydantic \texttt{BaseModel}, a typed, validated structure that
LLMs must conform to when generating output.

Listing~\ref{lst:schema} shows the abbreviated \texttt{GameOntology}
schema.  The \texttt{MechanismType} enumeration reflects ontological
categories developed over decades of game design
taxonomy~\cite{engelstein2020building}.  By encoding these as an enum, we
ensure the LLM can only reference mechanisms within our shared
vocabulary.  Every field in \texttt{GameOntology} corresponds to an
ontology concept.  The \texttt{goal} and \texttt{end\_condition} are
required strings with minimum lengths, preventing vague outputs.  The
nested \texttt{ComponentSet} and \texttt{PlayerDynamics} models enforce
structure at every level.

\begin{lstlisting}[caption={Abbreviated game ontology Pydantic schema.
Each field corresponds to an ontology concept; enums constrain the LLM
to recognized mechanisms.},label={lst:schema},float=*]
from enum import Enum
from typing import List, Literal, Optional
from pydantic import BaseModel, Field

class MechanismType(str, Enum):
    WORKER_PLACEMENT = "worker_placement"
    ACTION_POINTS    = "action_points"
    DECK_BUILDING    = "deck_building"
    RESOURCE_MGMT    = "resource_management"
    AREA_CONTROL     = "area_control"
    ENGINE_BUILDING  = "engine_building"
    SET_COLLECTION   = "set_collection"
    HIDDEN_INFO      = "hidden_information"

class GameOntology(BaseModel):
    title: str = Field(description="Name of the game")
    theme: str = Field(description="Setting and narrative")
    game_type: Literal["cooperative","competitive",
                        "semi-cooperative"]
    goal: str = Field(min_length=20,
        description="Victory condition")
    end_condition: str = Field(min_length=10,
        description="What triggers game end")
    primary_mechanisms: List[MechanismType] = Field(
        min_length=2, max_length=4)
    turn_structure: str
    uncertainty_source: str
    components: ComponentSet  # nested model
    players: PlayerDynamics   # nested model
    setup: str
    core_loop: str
    strategic_depth: str
\end{lstlisting}

\subsection{Ontology-to-Schema Correspondence}

Table~\ref{tab:correspondence} formalizes the mapping.  The
correspondence is not accidental: every ontology class finds expression
in the schema, and every ontological relationship (\emph{Game has
Mechanisms}, \emph{Game uses Components}) becomes a nested field or list.
The schema is the ontology made executable.

\begin{table}[t]
\centering
\caption{Ontology-to-schema correspondence.}
\label{tab:correspondence}
\footnotesize
\begin{tabular}{@{}lll@{}}
\toprule
\textbf{Concept} & \textbf{Schema Type} & \textbf{Purpose} \\
\midrule
Game Types & \texttt{Literal[...]} & Valid modes \\
Mechanisms & \texttt{List[Mechanism]} & Mechanics \\
Components & \texttt{ComponentSet} & Physical parts \\
Goal / End & \texttt{str} (required) & Playability \\
Players    & \texttt{PlayerDynamics} & Interactions \\
\bottomrule
\end{tabular}
\end{table}

\subsection{Why Pydantic?}

The choice of Pydantic~\cite{dspy2024docs} is deliberate.  Declaring
\texttt{GameOntology} as a \texttt{BaseModel} provides four critical
capabilities: (1)~\emph{type validation}, where the LLM's output is
automatically validated against the schema; (2)~\emph{self-documentation},
where each \texttt{Field} description becomes part of the prompt context;
(3)~\emph{nested structure}, where hierarchical ontologies map naturally
to nested models; and (4)~\emph{enum constraints}, where the LLM
operates within the ontological vocabulary and cannot invent mechanisms
outside the taxonomy.

% ============================================================
% 5. OPERATIONALIZING WITH DSPy
% ============================================================
\section{Operationalizing with DSPy}
\label{sec:dspy}

We have the ontology encoded as Pydantic schemas.  Now we need an engine
to drive generation.  DSPy (Declarative Self-improving
Python)~\cite{khattab2024dspy} treats LLM interactions as composable,
typed operations rather than monolithic prompt templates.  This
philosophy aligns with Generative Ontology: DSPy signatures \emph{are}
ontological contracts.

\subsection{Signatures as Ontological Contracts}

A DSPy signature declares inputs, outputs, and a docstring that
establishes the LLM's persona.  Listing~\ref{lst:signature} shows the
core generation signature.  The docstring becomes the system prompt; the
input fields capture what the user provides; the output field declares
that the result must conform to our \texttt{GameOntology} schema.

\begin{lstlisting}[caption={DSPy generation signature.  The typed output
field binds generation to the ontological schema.},label={lst:signature},float]
import dspy
from game_ontology import GameOntology

class GenerativeOntologySignature(dspy.Signature):
    """You are an expert tabletop game designer
    with deep knowledge of mechanisms, player
    dynamics, and component design. Design a
    complete, playable game based on the theme.
    Every field must be specific, not vague."""

    theme = dspy.InputField(
        desc="Thematic concept for the game")
    design_constraints = dspy.InputField(
        desc="Player count, complexity, time",
        default="")
    game_design: GameOntology = dspy.OutputField(
        desc="Complete game design per schema")
\end{lstlisting}

\subsection{Execution Flow}

When we execute this signature, DSPy performs the following sequence:
(1)~constructs a prompt incorporating the docstring, field descriptions,
and the full Pydantic schema structure; (2)~sends the prompt to the
configured LLM; (3)~parses the response into the \texttt{GameOntology}
model; (4)~validates that all required fields are present and correctly
typed; and (5)~returns the validated result or raises an error for retry.

DSPy's \texttt{ChainOfThought} module extends this by instructing the
model to reason through the design before committing to structured
output.  Encapsulating logic in a DSPy \texttt{Module} enables
composition of multiple generation steps, post-generation validation,
and retry logic.  The module adds a layer of semantic validation beyond
what Pydantic provides: the schema ensures structural correctness (all
fields present, correct types), while the module ensures
\emph{coherence}: that mechanisms match components and that game type
aligns with player dynamics.

% ============================================================
% 6. MULTI-AGENT ONTOLOGY PIPELINE
% ============================================================
\section{Multi-Agent Ontology Pipeline}
\label{sec:multiagent}

A single LLM call, even with our ontology schema, carries the entire
burden of game design: it must simultaneously consider mechanisms, theme,
components, balance, and player experience.  Human design teams do not
work this way.  Generative Ontology enables the same division of labor by
decomposing the ontology into domains and assigning specialized agents
to each.

\subsection{Agent Roster and Professional Anxieties}

Table~\ref{tab:agents} lists the five agents.  The ``Anxiety'' column is
crucial: each agent is not merely assigned a domain but is given a
\emph{concern}, a professional worry that shapes its generation and
critique.  This anxiety-driven design prevents the ``yes-man'' tendency
of LLMs to produce plausible but shallow output.

\begin{table*}[t]
\centering
\caption{Agent roster with ontology domains and professional anxieties.}
\label{tab:agents}
\small
\begin{tabular}{@{}lll@{}}
\toprule
\textbf{Agent} & \textbf{Domain} & \textbf{Anxiety} \\
\midrule
Mechanics Architect & Mechanisms, turn structure & ``Is there meaningful player agency?'' \\
Theme Weaver        & Narrative, setting         & ``Does the theme feel alive in every mechanism?'' \\
Component Designer  & Cards, tokens, board       & ``Can players manipulate this smoothly?'' \\
Balance Critic      & Cross-domain analysis      & ``What breaks when optimized?'' \\
Fun Factor Judge    & Player experience          & ``Would I want to play this again?'' \\
\bottomrule
\end{tabular}
\end{table*}

\begin{figure*}[t]
\centering
\includegraphics[width=0.95\textwidth]{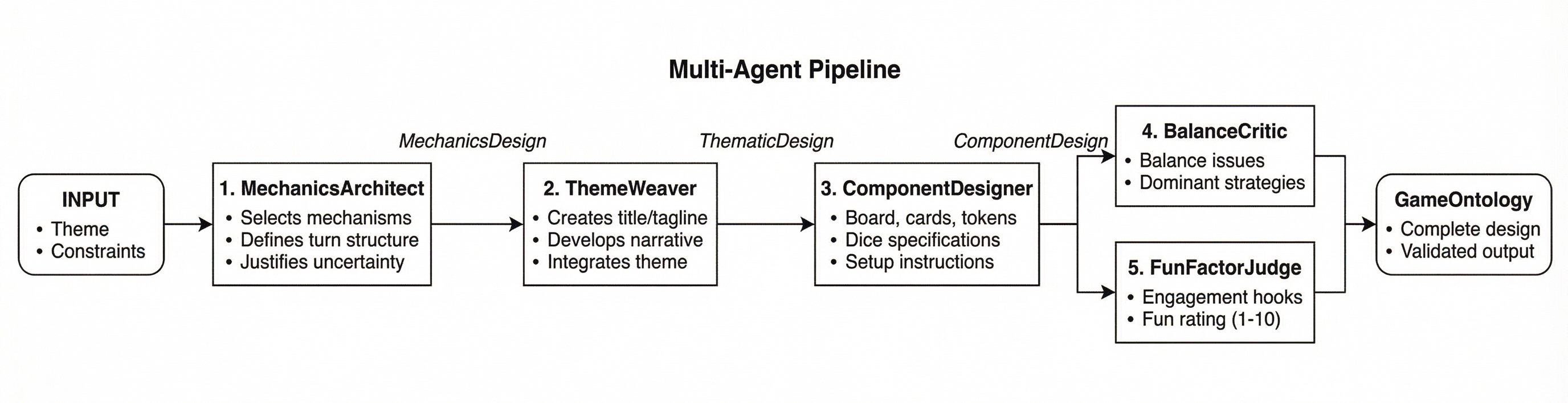}
\caption{Multi-agent pipeline architecture.  Generative agents build
sequentially (Mechanics $\to$ Theme $\to$ Components); critic agents
evaluate in parallel; refinement addresses identified issues before
final assessment.}
\label{fig:pipeline}
\end{figure*}

\subsection{Pipeline Architecture}

The pipeline, illustrated in Figure~\ref{fig:pipeline}, proceeds in four
phases:

\begin{enumerate}[nosep]
\item \textbf{Sequential Generation.}  The Mechanics Architect produces
  core systems given the theme and constraints.  The Theme Weaver
  receives these mechanics and integrates narrative.  The Component
  Designer receives both and specifies physical elements.
\item \textbf{Critical Evaluation.}  The Balance Critic examines the
  assembled design for exploits, dominant strategies, and interaction
  gaps.
\item \textbf{Refinement.}  If moderate or severe issues are found, a
  refinement agent addresses specific recommendations while preserving
  the design's strengths.
\item \textbf{Experience Assessment.}  The Fun Factor Judge evaluates
  engagement hooks, tension moments, satisfaction sources, and
  replayability.
\end{enumerate}

% ============================================================
% 7. RETRIEVAL-AUGMENTED GENERATION
% ============================================================
\section{Retrieval-Augmented Generation}
\label{sec:rag}

Our Generative Ontology can create novel game designs, but creativity
without context risks reinventing solutions that already exist.  The
most interesting creative work happens in conversation with what has
come before.

\begin{figure*}[t]
\centering
\includegraphics[width=0.65\textwidth]{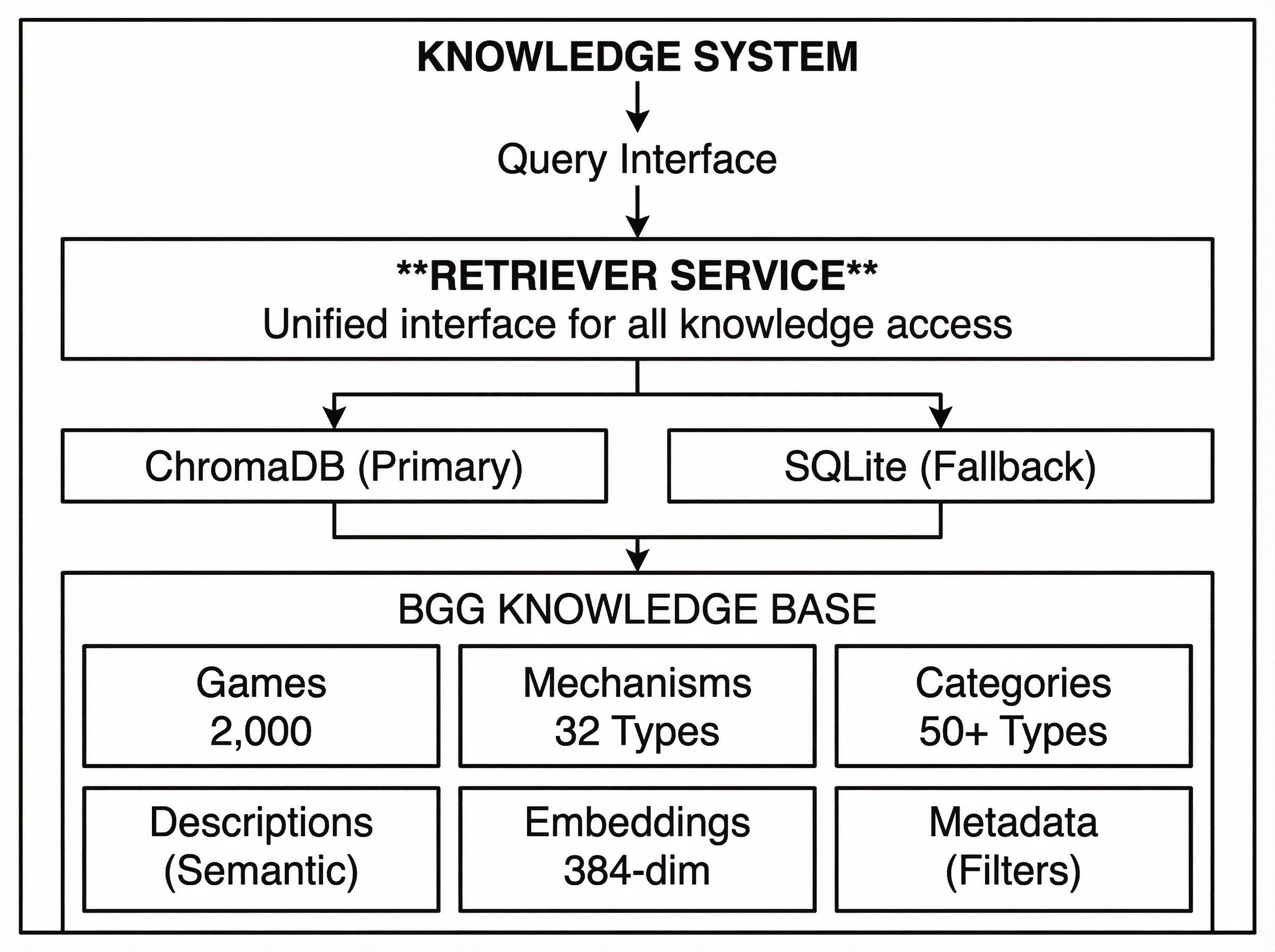}
\caption{Knowledge System architecture.  A unified Retriever Service
dispatches queries to ChromaDB (semantic vector search) or SQLite
(structured fallback), drawing on a BGG corpus of 1{,}767 games with
384-dimensional embeddings, mechanism types, and filterable metadata.}
\label{fig:rag}
\end{figure*}

\subsection{Two-Phase Retrieval}

BoardGameGeek~\cite{bgg} contains data on over 100,000 tabletop games.
We structure this corpus according to our ontology and employ a
two-phase retrieval strategy: (1)~\emph{ontology filtering}, which narrows
candidates to games sharing target mechanisms from the
\texttt{MechanismType} taxonomy; then (2)~\emph{semantic ranking}, which
ranks the filtered set by embedding similarity to the thematic query.
This ensures retrieved games are both \emph{mechanically relevant} and
\emph{thematically resonant}.

\subsection{Grounded Creativity}

RAG transforms Generative Ontology from invention to \emph{informed
invention}.  The system knows what mechanisms have been successfully
combined, how similar themes have been approached, and what design
patterns work for given player counts and complexity levels.  The
ontology provides the grammar; the retrieved examples provide the
vocabulary.  Together, they enable generation that speaks the language of
game design fluently.

% ============================================================
% 8. ONTOLOGY AS VALIDATION CONTRACT
% ============================================================
\section{Ontology as Validation Contract}
\label{sec:validation}

Generation is not enough.  The ontology serves not only as a generation
schema but as a validation contract: a set of constraints that generated
designs must satisfy to count as coherent.

\subsection{Ontological Constraint Checking}

We encode mechanism--component dependencies as validation functions.
Listing~\ref{lst:validator} shows the core checker: a dictionary maps
each mechanism to the component field it requires (deck building
$\to$~cards, area control $\to$~board, worker placement $\to$~tokens).
When a mechanism is present but its required component is absent, the
validator flags the incoherence.

\begin{lstlisting}[caption={Ontology validator excerpt checking
mechanism--component coherence.},label={lst:validator},float]
def _check_mechanism_requirements(
    self, design: GameOntology
) -> List[str]:
    issues = []
    requirements = {
        MechanismType.DECK_BUILDING:
            ("card_types",
             "Deck building needs cards"),
        MechanismType.AREA_CONTROL:
            ("board_description",
             "Area control needs board"),
        MechanismType.WORKER_PLACEMENT:
            ("tokens",
             "Workers need tokens"),
    }
    for mech in design.primary_mechanisms:
        if mech in requirements:
            field, msg = requirements[mech]
            val = getattr(
                design.components, field, None)
            if not val:
                issues.append(msg)
    return issues
\end{lstlisting}

\subsection{Iterative Refinement}

DSPy's \texttt{Assert} mechanism enables inline validation: when
\texttt{dspy.Assert(is\_valid, error\_message)} fails, DSPy
automatically retries generation with the error message included as
feedback.  This architecture treats the ontology as a \emph{contract}
between the generation system and downstream consumers.  Downstream
systems (whether human designers, balancing tools, or game
engines) can rely on structural guarantees: a deck-building game will
have cards; a cooperative game will not have direct conflict; the end
condition will be specified.

% ============================================================
% 9. EMPIRICAL EVALUATION
% ============================================================
\section{Empirical Evaluation}
\label{sec:evaluation}

The preceding sections describe the Generative Ontology framework.  We
now evaluate it through three complementary studies: an ablation study
isolating the contribution of each architectural component, a benchmark
comparison against published board games, and a test-retest reliability
analysis validating our evaluation instrument.

\subsection{Experimental Setup}
\label{sec:setup}

\noindent\textbf{Generation Model.}
All designs were generated using Claude Sonnet~4
(model ID \texttt{claude-sonnet-4-20250514},
temperature~0.7) via DSPy.

\noindent\textbf{Evaluation Pipeline.}
Each generated design is evaluated through two complementary lenses:
(1)~\emph{structural metrics}, computed deterministically by checking the
ontology for completeness (are all required fields present?), alignment
(do mechanisms match components?), and consistency (are there
contradictions?); and (2)~\emph{creative metrics}, scored by an LLM
evaluator (``Design Coach'') on seven dimensions: strategic depth,
tension/drama, player agency, replayability, social interaction,
elegance, and thematic cohesion, each on a 1--10 scale.  The Design
Coach also provides an overall fun rating (1--10) and engagement
variance (standard deviation across dimension scores).  The evaluator
uses the same model at temperature 0 for near-deterministic scoring.

\noindent\textbf{Theme Prompts.}
Ten standardized theme prompts span a range of complexity levels and
game types (see Appendix~\ref{app:prompts}).  Each prompt specifies a
theme, player count, complexity level, and target play time.

\noindent\textbf{Experimental Conditions.}
Four conditions isolate the contribution of each pipeline component
(Table~\ref{tab:conditions}).

\begin{table}[t]
\centering
\caption{Experimental conditions for the ablation study.}
\label{tab:conditions}
\small
\begin{tabularx}{\columnwidth}{lX}
\toprule
\textbf{Condition} & \textbf{Configuration} \\
\midrule
C1 (Baseline) & Raw LLM generation, no schema or ontology \\
C2 (Schema) & Pydantic schema validation only \\
C3 (Ontology) & Schema + ontology constraints, single agent \\
C4 (Pipeline) & Full pipeline: schema + ontology + RAG + multi-agent + validation \\
\bottomrule
\end{tabularx}
\end{table}

% ---------- Study 1: Ablation ----------
\subsection{Study 1: Ablation}
\label{sec:ablation}

\noindent\textbf{Design.}
$4 \text{ conditions} \times 10 \text{ prompts} \times 3
\text{ replications} = 120$ designs, all evaluated through the full
pipeline.

\noindent\textbf{Structural Validity.}
Table~\ref{tab:structural} reports structural metrics.  Schema
validation (C1$\to$C2) produces the largest single improvement: mean
consistency errors drop from 5.03 to 0.10 ($d = 4.78$, $p < .001$).
Adding ontology constraints (C3, C4) eliminates errors entirely.  C3
achieves perfect consistency but lower completeness (83\%) because
single-agent generation misses some detail fields, resolved by
multi-agent specialization in C4.

\begin{table}[t]
\centering
\caption{Structural metrics by condition.}
\label{tab:structural}
\small
\begin{tabular}{@{}lccc@{}}
\toprule
\textbf{Condition} & \textbf{Errors} & \textbf{Compl.} \\
\midrule
C1 (Baseline) & 5.03 & 100\% \\
C2 (Schema)   & 0.10 & 100\% \\
C3 (Ontology) & 0.00 &  83\% \\
C4 (Pipeline) & 0.00 & 100\% \\
\bottomrule
\end{tabular}
\end{table}

\begin{figure*}[t]
\centering
\includegraphics[width=0.95\textwidth]{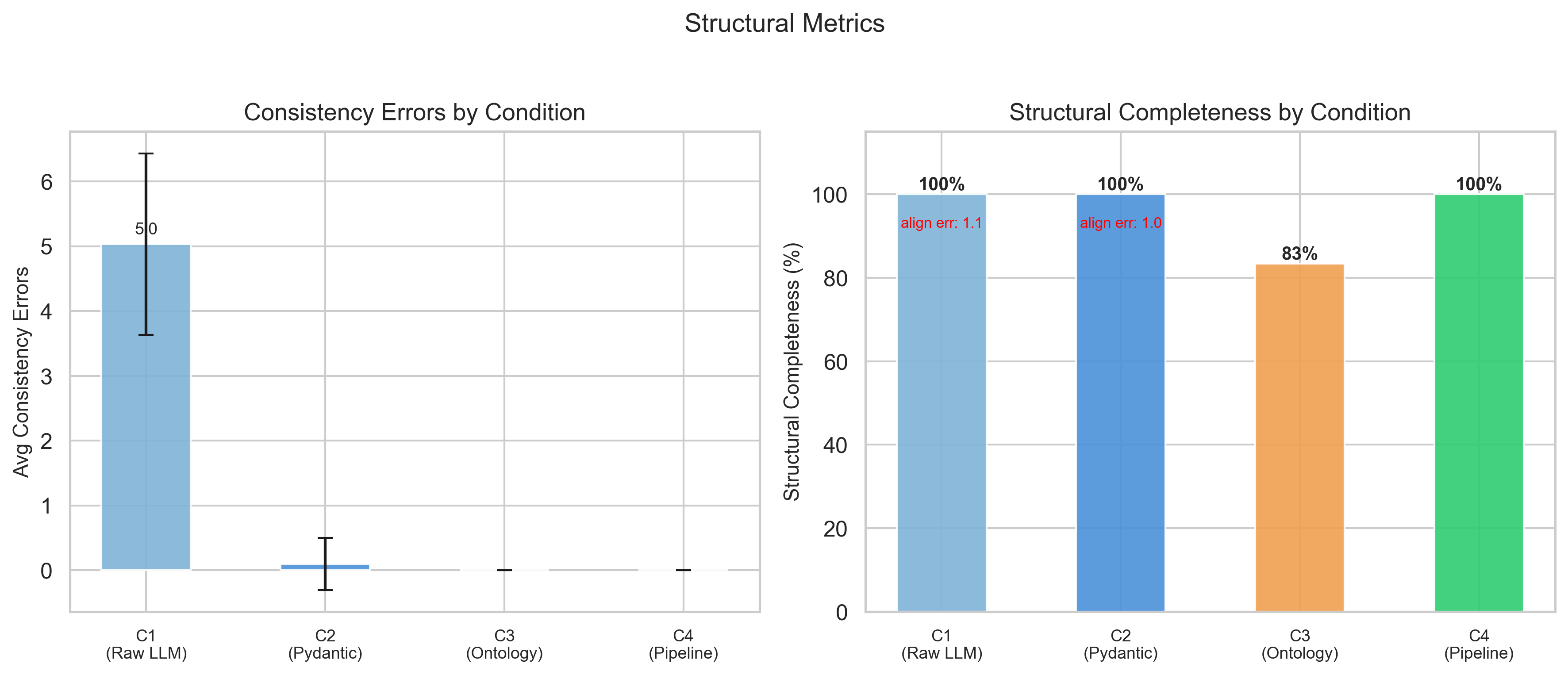}
\caption{Structural metrics by condition.  Left: mean consistency errors
drop from 5.03 (C1) to zero (C3, C4).  Right: completeness is 100\%
for all conditions except C3 (83\%), where single-agent generation
misses detail fields.}
\label{fig:structural}
\end{figure*}

\begin{figure*}[t]
\centering
\includegraphics[width=0.95\textwidth]{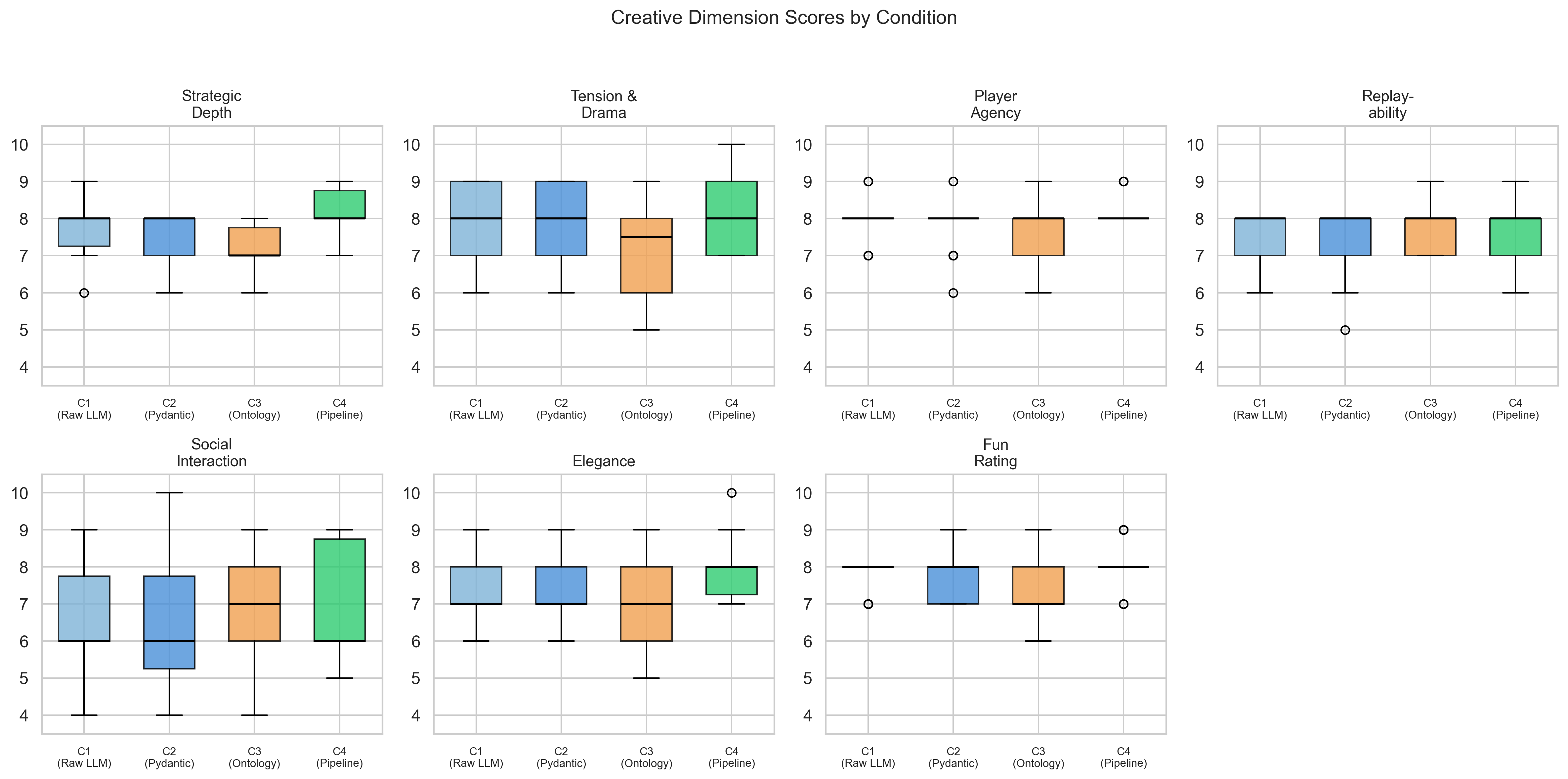}
\caption{Creative dimension score distributions by condition.  C4
(Pipeline) shows the highest and tightest distributions on most
dimensions.  Replayability and social interaction do not discriminate
between conditions.}
\label{fig:boxplots}
\end{figure*}

\noindent\textbf{Creative Quality.}
Figure~\ref{fig:boxplots} shows the distribution of creative dimension
scores.  The key finding is that constraints alone (C2, C3) do not
improve creative quality; multi-agent specialization (C4) does.  The
largest effects from C3$\to$C4 are on strategic depth ($d = 1.59$,
$p < .001$), elegance ($d = 1.14$, $p < .001$), and fun ($d = 1.12$,
$p < .001$).  Tension shows a moderate effect ($d = 0.79$, $p < .05$).
Replayability and social interaction show no reliable differences across
any conditions.

\begin{figure}[t]
\centering
\includegraphics[width=\columnwidth]{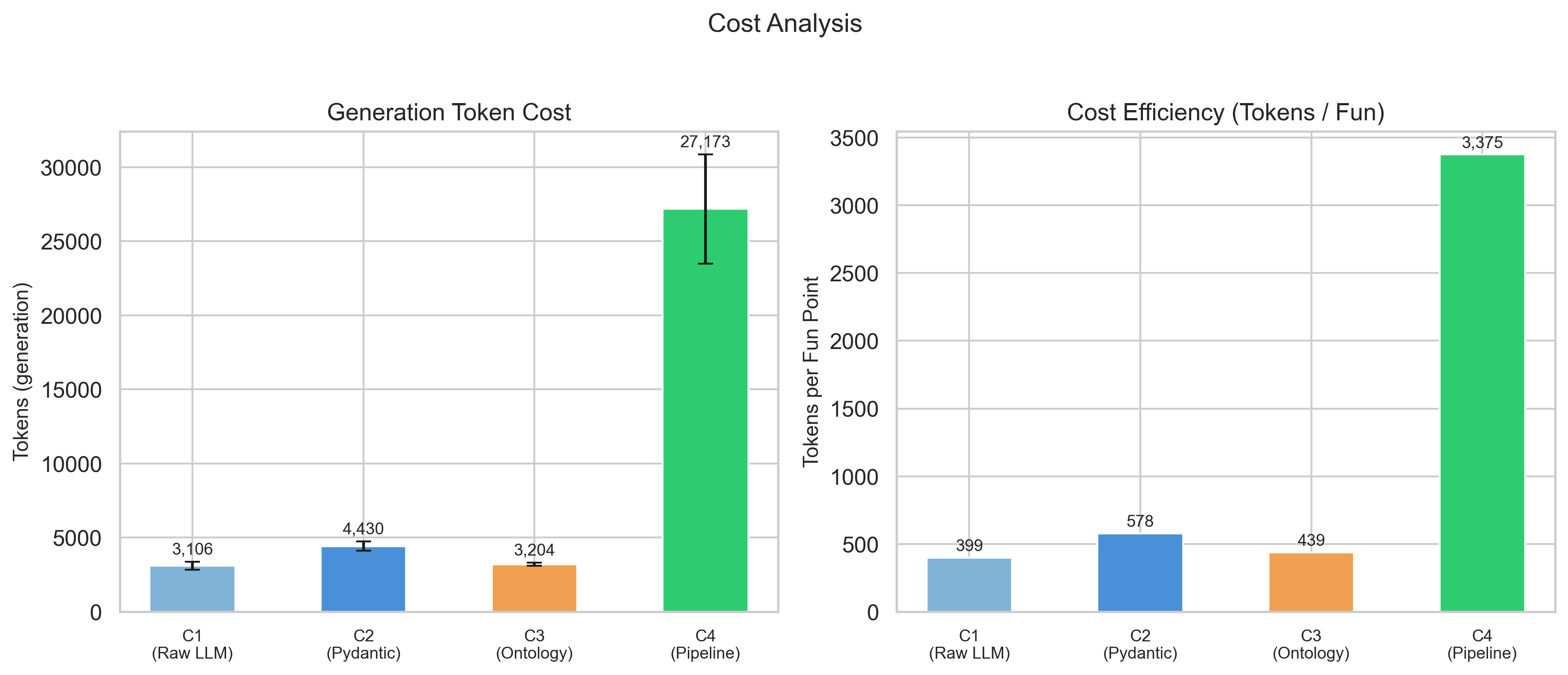}
\caption{Cost analysis.  Left: generation tokens per design.
Right: tokens per fun point.  C4 costs ${\sim}8.5\times$ more but
achieves the highest fun rating.}
\label{fig:cost}
\end{figure}

\noindent\textbf{Cost--Quality Frontier.}
The multi-agent pipeline incurs significant computational cost.  C4
uses ${\sim}27$k tokens per design versus ${\sim}3.1$k for C1, an
8.5$\times$ increase.  Figure~\ref{fig:cost} shows both absolute cost
and the efficiency ratio (tokens per fun point).  C4 achieves the
highest quality but at 3{,}369 tokens per fun point versus 398 for C1,
establishing a clear Pareto frontier.

\noindent\textbf{Summary.}
Structural validity is solved by schema validation.  Creative quality
arises from the interaction between constraint enforcement and
architectural specialization: constraints alone suppress richness;
specialization restores and enhances it.

% ---------- Study 2: Benchmark ----------
\subsection{Study 2: Benchmark Comparison}
\label{sec:benchmark}

\noindent\textbf{Design.}
We compare 30 C4-generated designs from Study~1 against 20 published
board games evaluated through the same pipeline.  The reference set
spans three complexity tiers: light (Azul, Splendor, Love Letter,
Kingdomino, Codenames), medium (Wingspan, Everdell, 7~Wonders,
Concordia, Spirit Island), and heavy (Brass: Birmingham, Gaia Project,
Twilight Imperium 4E, Through the Ages, Gloomhaven), plus five
pre-existing ontologies (Catan, Root, Dune Imperium, Terraforming Mars,
Ticket to Ride).  See Appendix~\ref{app:games} for the full list.
Each real game was reverse-engineered into a complete GameGrammar
ontology and evaluated through the identical Design Coach pipeline.

\noindent\textbf{Structural Parity.}
Both groups achieve 100\% completeness.  Generated designs have fewer
consistency errors (1.27 vs.~2.80, $d = 0.76$, $p < .01$).  The higher
error count for real games reflects richer component vocabularies
(specialized tokens, standees, custom boards) that extend beyond the
checker's standard categories.

\noindent\textbf{Creative Gap.}
Table~\ref{tab:benchmark} reports creative dimension scores.  Published
games score significantly higher on 5 of 7 dimensions, with the largest
effects on elegance ($d = 1.72$) and fun ($d = 1.86$).  Crucially,
tension/drama ($d = 0.35$, ns) and social interaction ($d = 0.19$, ns)
show no significant differences, suggesting the pipeline already handles these
dimensions comparably.

\begin{table}[t]
\centering
\caption{Creative dimensions: real vs.\ generated designs.}
\label{tab:benchmark}
\footnotesize
\begin{tabular}{@{}lcccc@{}}
\toprule
\textbf{Dimension} & \textbf{Real} & \textbf{Gen.} & \textbf{$d$} & \textbf{Sig.} \\
\midrule
Strategic Depth   & 8.85 & 8.13 & 0.94  & **  \\
Tension \& Drama  & 8.50 & 8.20 & 0.35  & ns  \\
Player Agency     & 9.00 & 8.23 & 1.21  & *** \\
Replayability     & 9.05 & 7.63 & 1.59  & *** \\
Social Interaction& 7.20 & 6.93 & 0.19  & ns  \\
Elegance          & 9.25 & 7.97 & 1.72  & *** \\
Fun Rating        & 8.90 & 8.07 & 1.86  & *** \\
\bottomrule
\end{tabular}

\medskip
\raggedright\scriptsize
$d$ = Cohen's $d$; ns = not significant; ** $p < .01$; *** $p < .001$.
\end{table}

\begin{figure*}[t]
\centering
\includegraphics[width=0.95\textwidth]{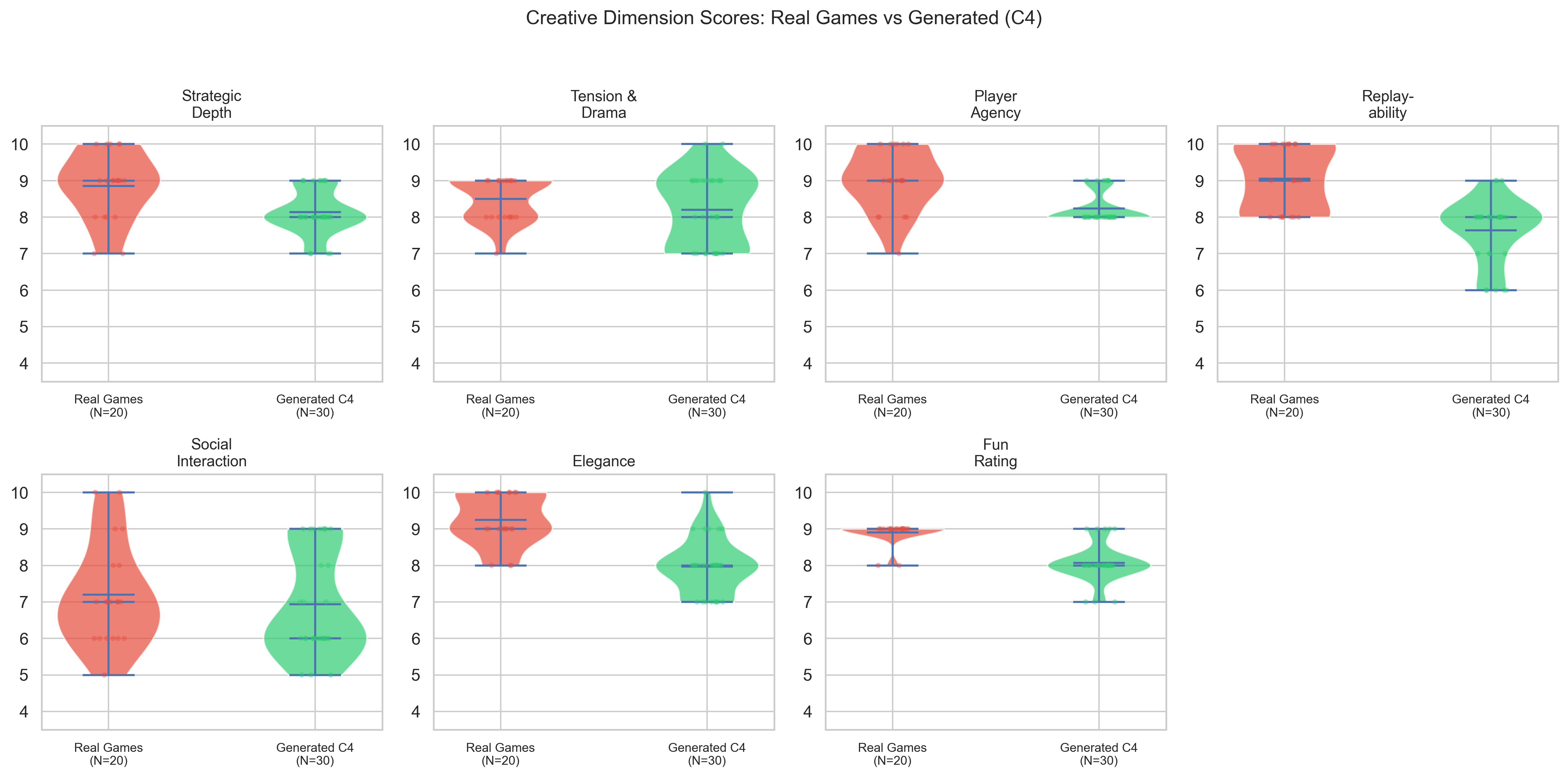}
\caption{Violin plots showing score distributions for all creative
dimensions.  Real games (red) cluster in the 8--9 range while generated
designs (green) spread across 7--9.  Tension/drama and social interaction
show the most overlap.}
\label{fig:violins}
\end{figure*}

\begin{figure}[t]
\centering
\includegraphics[width=\columnwidth]{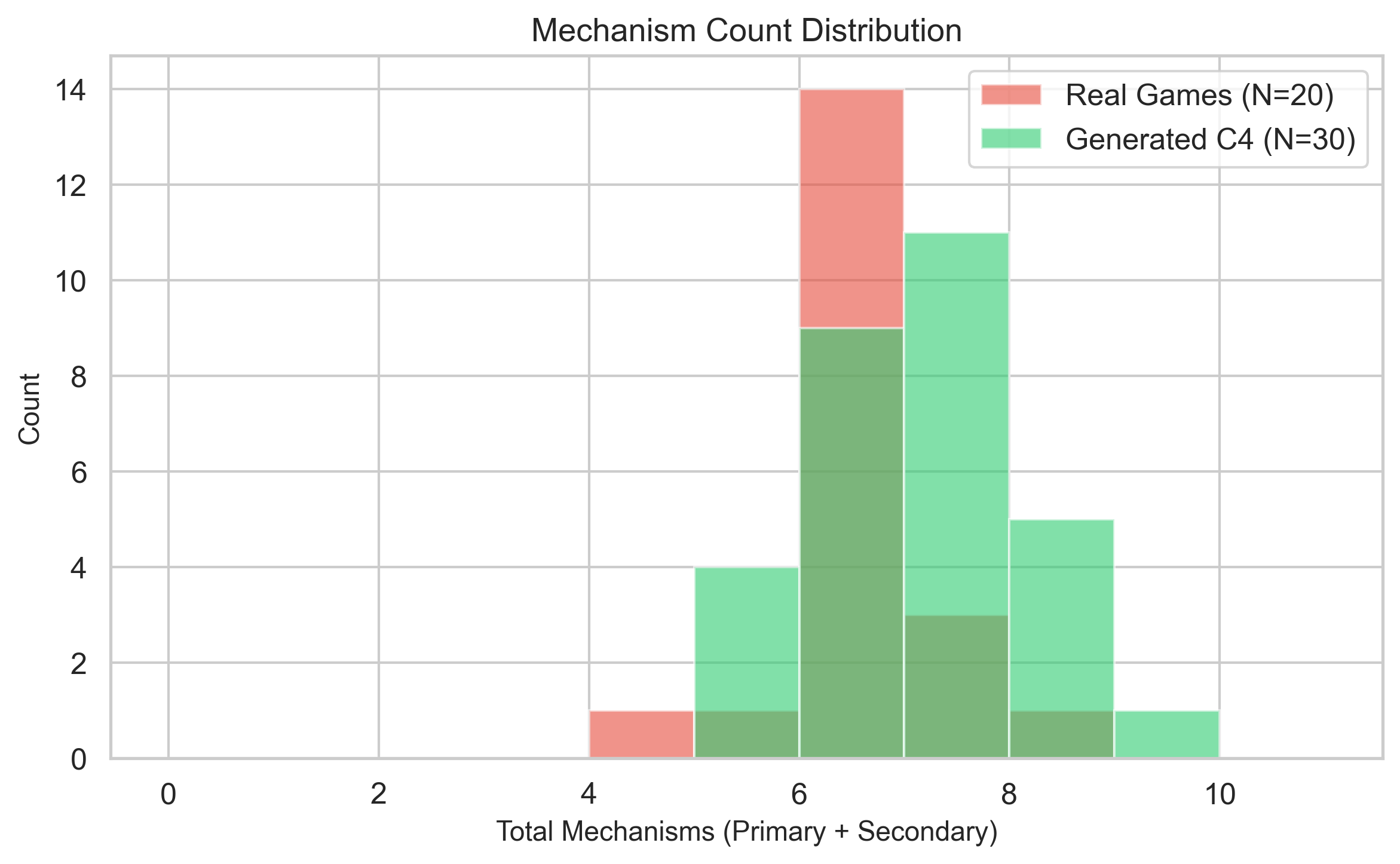}
\caption{Mechanism count distributions.  Both groups center around
6--7 mechanisms; generated designs show slightly more spread.}
\label{fig:mechanisms}
\end{figure}

\noindent\textbf{Mechanism Similarity.}
Generated designs use a similar number of mechanisms (6.7 vs.~6.1,
$d = -0.60$) but fewer component types (7.8 vs.~11.6, $d = 1.12$,
$p < .01$).  The pipeline generates designs within a mid-complexity band
but does not yet produce the extremes: neither ultra-light party games
nor epic-scale strategy games (Figure~\ref{fig:mechanisms}).

\noindent\textbf{Summary.}
Generated designs achieve structural parity with published games but
exhibit a bounded creative gap of approximately one point on a 10-point
scale.  This gap likely reflects properties that emerge from iterative
playtesting (elegance and replayability) rather than design
specification alone.

% ---------- Study 3: Reliability ----------
\subsection{Study 3: Evaluator Reliability}
\label{sec:reliability}

An empirical evaluation is only as credible as its evaluation instrument.
We assess the Design Coach's test-retest reliability using intraclass
correlation coefficients (ICC).

\noindent\textbf{Design.}
Ten designs spanning the quality range (fun ratings 6--9, all four
conditions) were each evaluated five times through the Design Coach,
yielding 50 evaluations.  We compute ICC(2,1), two-way random effects with
single rater and absolute agreement~\cite{shrout1979intraclass}, for each
of the nine LLM-scored metrics.

\noindent\textbf{Results.}
Table~\ref{tab:icc} reports ICC values.  Seven of nine metrics achieve
Good-to-Excellent reliability (ICC $\geq 0.75$).  Four metrics rate
Excellent ($\geq 0.90$): tension \& drama (0.989), social interaction
(0.983), thematic cohesion (0.975), and elegance (0.934).  Three rate
Good: fun rating (0.873), strategic depth (0.846), and player agency
(0.836).  Two metrics, replayability (0.578) and engagement variance
(0.598), show moderate reliability.

\begin{table}[t]
\centering
\caption{Test-retest reliability: ICC(2,1) per metric.}
\label{tab:icc}
\footnotesize
\begin{tabular}{@{}lccl@{}}
\toprule
\textbf{Metric} & \textbf{ICC} & \textbf{95\% CI} & \textbf{Level} \\
\midrule
Tension \& Drama    & .989 & [.970, 1.00] & Excellent \\
Social Interaction  & .983 & [.960, .990] & Excellent \\
Thematic Cohesion   & .975 & [.940, .990] & Excellent \\
Elegance            & .934 & [.850, .980] & Excellent \\
Fun Rating          & .873 & [.730, .960] & Good \\
Strategic Depth     & .846 & [.680, .950] & Good \\
Player Agency       & .836 & [.660, .950] & Good \\
Engagement Var.     & .598 & [.330, .850] & Moderate \\
Replayability       & .578 & [.300, .840] & Moderate \\
\bottomrule
\end{tabular}
\end{table}

\begin{figure}[t]
\centering
\includegraphics[width=\columnwidth]{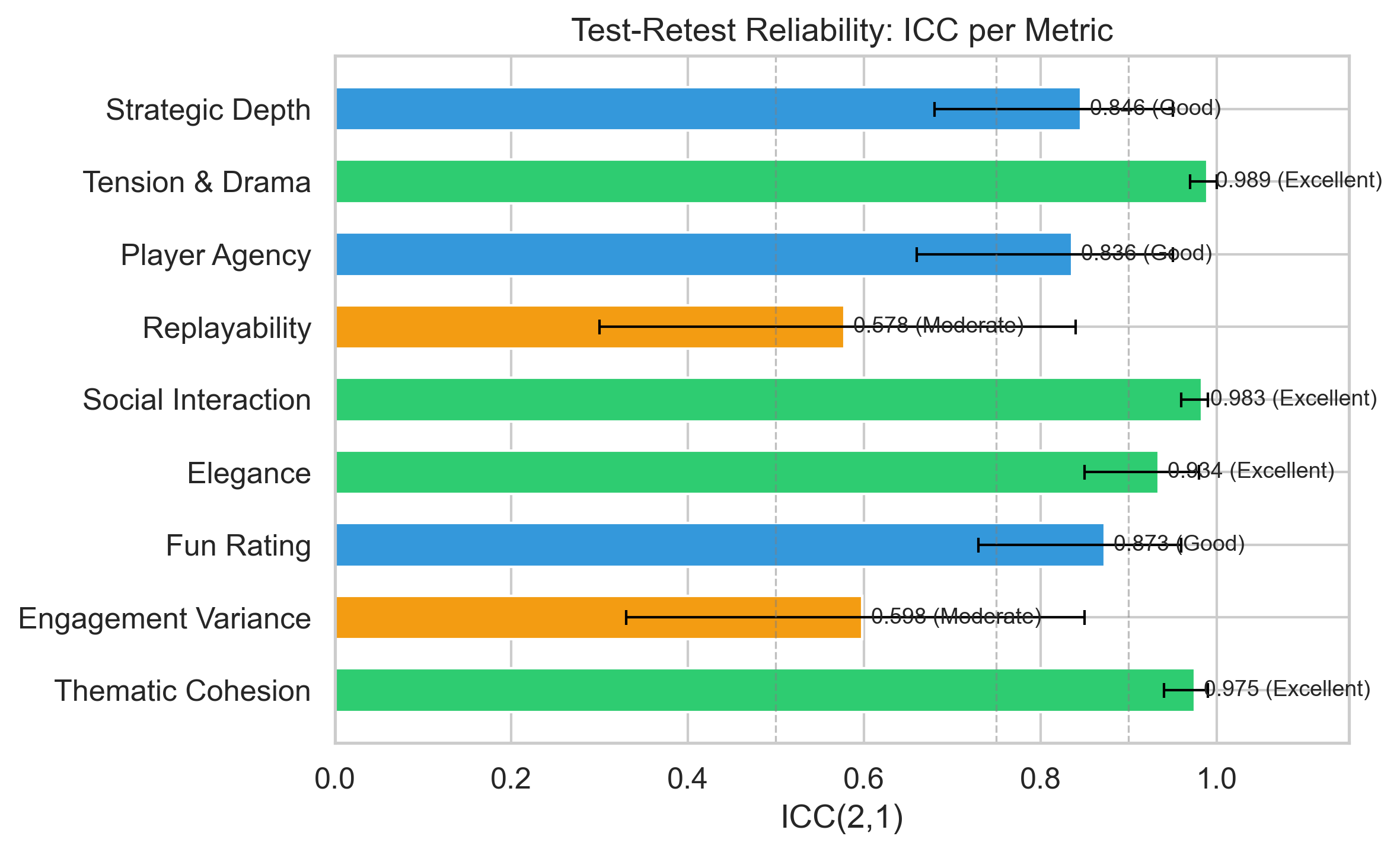}
\caption{ICC values per metric with 95\% confidence intervals.
Green = Excellent ($\geq .90$), blue = Good (.75--.90),
orange = Moderate (.50--.75).}
\label{fig:icc}
\end{figure}

\noindent\textbf{Cross-Study Validation.}
The ICC results retroactively validate prior findings.  Fun rating
(ICC = 0.873) confirms that the fun-based comparisons in Studies~1
and~2 rest on stable measurements.  The high ICC for tension \& drama
(0.989) and social interaction (0.983) confirms that the
non-significant differences in Study~2 reflect genuine equivalence, not
measurement noise.  Conversely, replayability's moderate ICC (0.578)
suggests that its non-significance in Study~1 may partly reflect
evaluator inconsistency, a property arguably better assessed through
playtesting than design specification analysis.

\noindent\textbf{Summary.}
The Design Coach is a reliable evaluation instrument.  Score differences
between designs reflect genuine quality differences, not evaluator drift.

% ============================================================
% 10. CASE STUDY: MYCELIUM: THE DEEP
% ============================================================
\section{Case Study: Mycelium: The Deep}
\label{sec:casestudy}

To illustrate the pipeline concretely, we trace a single generation
from theme to playable design.  The input is a theme chosen to be
unusual enough that retrieval finds no direct precedents:

\begin{quote}
\textbf{Theme:} ``Bioluminescent fungi competing for dominance in a deep
cave ecosystem''\\
\textbf{Constraints:} 2--4 players, medium complexity, cooperative,
60--90 minutes
\end{quote}

The Mechanics Architect selects three core mechanisms: resource
management (spore allocation), area control (cave network expansion),
and engine building (specialized fungal structures).  The Theme Weaver
titles the game \emph{Mycelium: The Deep} and maps each mechanism to its
fictional counterpart: spore placement as reproductive strategy, area
control as securing nutrient pathways, engine building as developing
specialized structures.  The Component Designer specifies 30 modular
hex cave tiles, three card types, and tokens including fungal growth
cubes, spore tokens, and flood markers.  The Balance Critic identifies a
runaway-leader problem and arbitrary flood timing; the refinement agent
adds a Network Synergy rule and predictable flood escalation.  The Fun
Factor Judge rates the refined design 7/10.  Table~\ref{tab:output}
summarizes the output.

\begin{table}[t]
\centering
\caption{Abbreviated output: \emph{Mycelium: The Deep}.}
\label{tab:output}
\small
\begin{tabularx}{\columnwidth}{lX}
\toprule
\textbf{Field} & \textbf{Value} \\
\midrule
Title & Mycelium: The Deep \\
Type & Cooperative \\
Goal & Grow fungal network from cave depths to surface before flood reaches highest colony \\
End & Victory: growth cube reaches surface tile.  Defeat: flood reaches highest cube row \\
Mechanisms & Resource management, area control, engine building \\
Turn loop & Place spores $\to$ Reveal $\to$ Grow $\to$ Event $\to$ Flood rises \\
Players & 2--4, role-differentiated \\
Components & 30 hex tiles, 3 card types, 60 growth cubes, 12 spore tokens, 20 flood markers \\
\bottomrule
\end{tabularx}
\end{table}

Without the ontology schema, an LLM generating ``a cooperative game
about bioluminescent fungi'' would likely produce vague victory
conditions, inconsistent mechanism--component mappings, and disconnected
theme.  Generative Ontology enforced complete goal specification,
coherent mechanism--component alignment, and thematic integration.  The
output is not merely plausible; it is \emph{playable}.

% ============================================================
% 11. GENERALIZABILITY
% ============================================================
\section{Generalizability}
\label{sec:generalizability}

We developed Generative Ontology through tabletop games because games
have rich ontological structure and connect to our previous work.  But
the pattern extends to any domain with coherent structure, where experts
share vocabulary, where valid outputs satisfy definable constraints, and
where knowledge accumulates in exemplars.

\begin{figure*}[t]
\centering
\includegraphics[width=0.85\textwidth]{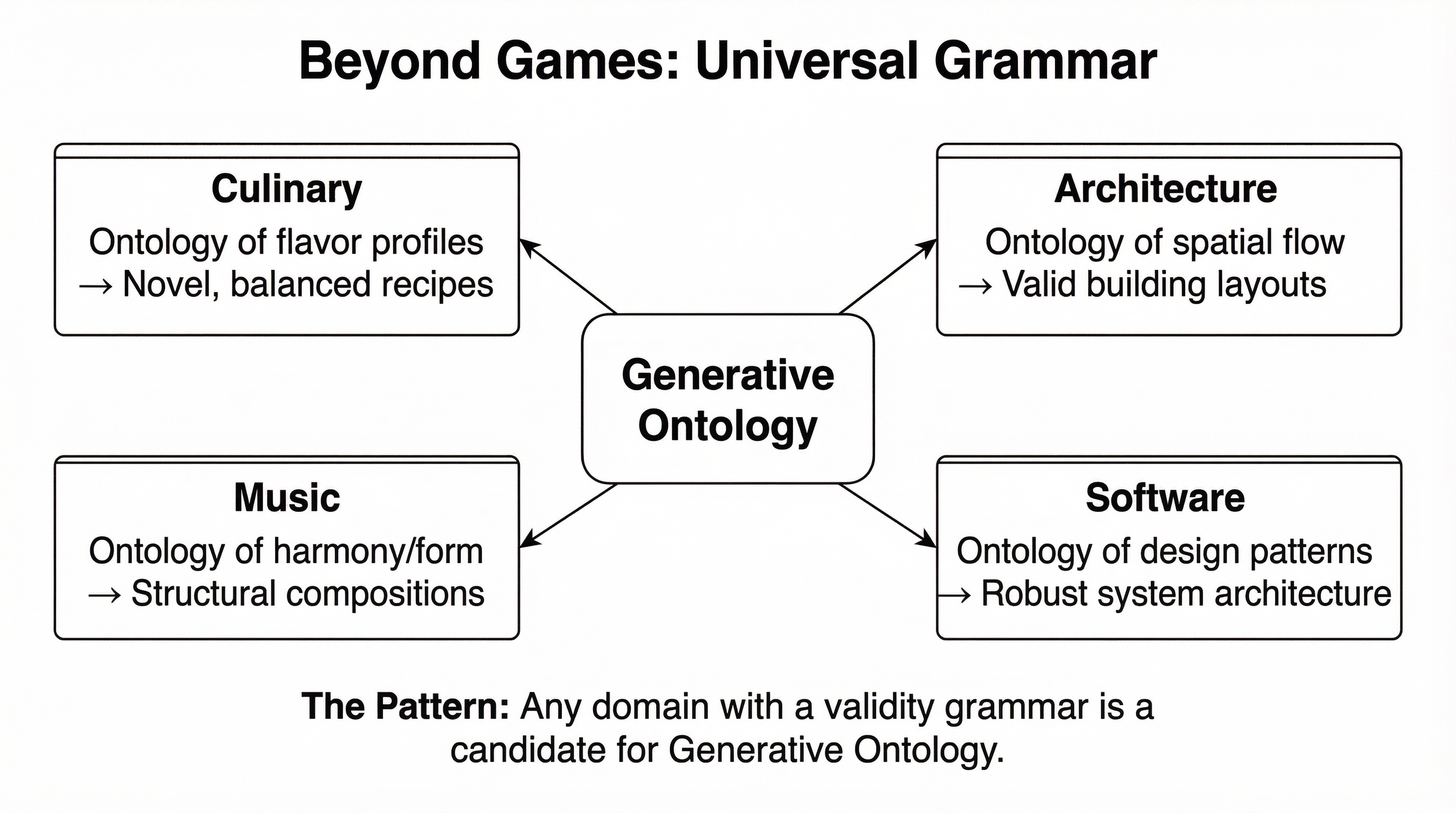}
\caption{Beyond games: any domain with a ``validity grammar'' is a
candidate for Generative Ontology.}
\label{fig:universal}
\end{figure*}

Table~\ref{tab:domains} illustrates eight domains where Generative
Ontology applies.  The pattern is most valuable when four conditions
hold: (1)~\emph{the domain has structure}: experts can articulate what
makes outputs valid; (2)~\emph{exemplars exist}: a corpus of successful
examples enables RAG; (3)~\emph{validity is checkable}: constraints can
be encoded as validation functions; and (4)~\emph{generation is
valuable}: the effort of building the ontology is justified by repeated
generation needs.

\begin{table*}[t]
\centering
\caption{Cross-domain applicability of Generative Ontology.}
\label{tab:domains}
\small
\begin{tabularx}{\textwidth}{l>{\raggedright\arraybackslash}X>{\raggedright\arraybackslash}X>{\raggedright\arraybackslash}X}
\toprule
\textbf{Domain} & \textbf{Ontology Elements} & \textbf{Generation Output} & \textbf{Validation Criteria} \\
\midrule
Tabletop Games & Mechanisms, components, dynamics & Novel game designs & Playability, balance, coherence \\
Music Composition & Scales, progressions, forms & Original compositions & Harmonic validity, structure \\
Architecture & Spaces, materials, flows, codes & Building designs & Structural integrity, compliance \\
Culinary Arts & Ingredients, techniques, flavors & Novel recipes & Flavor balance, feasibility \\
Narrative & Characters, arcs, themes & Stories and plots & Narrative coherence, pacing \\
Software Design & Patterns, components, interfaces & System architectures & Consistency, coupling, coverage \\
Legal Documents & Clauses, parties, obligations & Contracts & Legal validity, completeness \\
Scientific Protocols & Methods, materials, controls & Experimental designs & Reproducibility, validity \\
\bottomrule
\end{tabularx}
\end{table*}

% ============================================================
% 12. DISCUSSION
% ============================================================
\section{Discussion}
\label{sec:discussion}

\subsection{The Constraint Paradox in Data Form}

Our ablation results provide empirical evidence for what we term the
\emph{constraint paradox}: adding constraints (C2, C3) eliminates
structural errors without improving creative quality, yet the
combination of constraints with architectural specialization (C4)
produces the largest creative gains.  Constraints alone suppress
richness; constraints plus specialization enable it.  This suggests a
\emph{Constraint--Architecture Interaction Model}: creative quality
is a function of constraint expressiveness \emph{multiplied by}
architectural specialization, not simply additive.

\subsection{The Gap as Feature}

The one-point creative gap between generated and published designs
(Study~2) is, paradoxically, encouraging.  Published games represent
decades of iterative playtesting, community feedback, and designer
refinement, processes that no single generation pass can replicate.
That the pipeline produces designs in the 7--8 range (``good, playable,
interesting'') rather than the 8--9 range (``polished, elegant,
replayable'') suggests the gap reflects not a fundamental limitation
of the approach but the absence of iterative refinement.  The
dimensions where parity is already achieved (tension/drama and social
interaction) may be those most amenable to specification-time design,
while the gap dimensions (elegance and replayability) may require
iterative playtesting to optimize.

\subsection{The Whiteheadian Connection}

Generative Ontology offers a concrete instantiation of Whitehead's
metaphysics~\cite{whitehead1929process,cheung2026process}.  The ontology
provides eternal objects, abstract patterns existing as potentials.
Generation produces actual occasions, concrete instantiations where
form meets matter.  Creativity is the movement from potentiality to
actuality.  Our empirical results show this movement is not random: the
constraints shape the space of possible actualizations, and
architectural specialization navigates that space effectively.

\subsection{Limitations}

Several limitations qualify our findings.
\textbf{LLM-as-evaluator:}  While our ICC analysis demonstrates
evaluator reliability, it does not establish validity: LLM scores may
be consistently biased.  Human calibration (comparing LLM scores to
expert judgment) remains future work.
\textbf{No playtesting:}  Our evaluation assesses design specifications,
not gameplay.  Properties like elegance and replayability may manifest
differently in play than in specification.
\textbf{Reference set:}  The 20 benchmark games, while spanning three
complexity tiers, represent a convenience sample of well-known titles.
A larger, more diverse reference set would strengthen generalizability.
\textbf{Single model:}  All generation and evaluation used Claude
Sonnet~4.  Cross-model evaluation would assess robustness.
\textbf{Temperature sensitivity:}  Evaluator ICC was measured at
temperature~0; reliability at higher temperatures is untested.

% ============================================================
% 13. CONCLUSION AND FUTURE WORK
% ============================================================
\section{Conclusion and Future Work}
\label{sec:conclusion}

We began with a question: what happens when ontology learns to create?
The answer is Generative Ontology, a synthesis combining the structural
precision of knowledge representation with the creative power of large
language models~\cite{mei2025context}.

Four insights emerged from this work:

\noindent\textbf{The Grammar Paradox.}
Constraints enable creativity.  Our ablation study confirms this
empirically: schema validation eliminates structural errors ($d = 4.78$)
while multi-agent specialization produces the largest creative gains
(fun $d = 1.12$, depth $d = 1.59$).

\noindent\textbf{Neither Alone Suffices.}
Traditional ontology without generation is static.  LLM generation
without ontology is unconstrained.  The benchmark comparison shows that
the synthesis produces designs that are structurally equivalent to
published games, with a bounded creative gap that likely reflects the
absence of iterative playtesting rather than fundamental architectural
limitations.

\noindent\textbf{Reliable Measurement Matters.}
Our test-retest analysis (ICC 0.578--0.989, median 0.873) demonstrates
that LLM-based evaluation can be reliable when properly constrained.
The cross-study validation differentiates ``non-significant due to
noise'' from ``non-significant due to genuine equivalence,'' a
methodological contribution applicable beyond game design.

\noindent\textbf{The Pattern Generalizes.}
Although we developed Generative Ontology through games, the pattern
applies wherever structured domains exist.  Music, architecture, cooking,
narrative, software design: each has its grammar, its ontology, its
space of valid outputs.

\medskip

Future directions include \emph{human calibration} (validating LLM
scores against expert judgment), \emph{iterative design loops} (multiple
generation--evaluation--refinement cycles to close the creative gap),
\emph{cross-model evaluation} (assessing robustness across LLMs), and
\emph{DSPy optimization} (using the ontology validation function as a
reward signal for automatic prompt tuning).

\medskip

The grammar does not write the poem.  But without grammar, there is no
poem to write.  Generative Ontology gives AI the grammar it needs to
create meaningfully within any structured domain.  The
\textsc{GameGrammar} system is available at
\url{https://github.com/bennycheung/GameGrammarCLI}.

% ============================================================
% BIBLIOGRAPHY
% ============================================================
{\raggedright
\bibliographystyle{plainnat}
\bibliography{refs}

\begin{thebibliography}{20}
\providecommand{\natexlab}[1]{#1}
\providecommand{\url}[1]{\texttt{#1}}
\expandafter\ifx\csname urlstyle\endcsname\relax
  \providecommand{\doi}[1]{doi: #1}\else
  \providecommand{\doi}{doi: \begingroup \urlstyle{rm}\Url}\fi

\bibitem[bgg(2024)]{bgg}
{BoardGameGeek}.
\newblock \url{https://boardgamegeek.com/}, 2024.

\bibitem[dsp(2024)]{dspy2024docs}
{DSPy}: Programming with foundation models.
\newblock \url{https://dspy.ai/}, 2024.

\bibitem[Barker(2024)]{barker2024artificial}
Timothy Barker.
\newblock Artificial creativity: A process philosophy of technology
  perspective.
\newblock \emph{Journal of Continental Philosophy}, 2024.
\newblock URL \url{https://eprints.gla.ac.uk/327708/1/327708.pdf}.

\bibitem[Becker et~al.(2025)Becker, Oliveira, Rossato, and
  Tavares]{becker2025boardwalk}
{\'A}lvaro~Guglielmin Becker, Gabriel Bauer~de Oliveira, Lana~Bertoldo Rossato,
  and Anderson~Rocha Tavares.
\newblock Boardwalk: Towards a framework for creating board games with {LLMs}.
\newblock \emph{arXiv preprint arXiv:2508.16447}, 2025.
\newblock URL \url{https://arxiv.org/abs/2508.16447}.

\bibitem[Cheung(2025)]{cheung2025tabletop}
Benny Cheung.
\newblock Unlocking the secrets of tabletop games ontology.
\newblock
  \url{https://bennycheung.github.io/unlocking-secrets-of-tabletop-games-ontology},
  2025.

\bibitem[Cheung(2026)]{cheung2026process}
Benny Cheung.
\newblock Process philosophy for {AI} agent design.
\newblock
  \url{https://bennycheung.github.io/process-philosophy-for-ai-agent-design},
  2026.

\bibitem[Cooper(2024)]{cooper2024constrained}
Aidan Cooper.
\newblock A guide to structured generation using constrained decoding.
\newblock \url{https://www.aidancooper.co.uk/constrained-decoding/}, 2024.

\bibitem[Engelstein and Shalev(2020)]{engelstein2020building}
Geoffrey Engelstein and Isaac Shalev.
\newblock \emph{Building Blocks of Tabletop Game Design: An Encyclopedia of
  Mechanisms}.
\newblock CRC Press, 2020.

\bibitem[Gallotta et~al.(2024)]{gallotta2024llm}
Roberto Gallotta et~al.
\newblock Large language models and games: A survey and roadmap.
\newblock \emph{arXiv preprint arXiv:2402.18659}, 2024.
\newblock URL \url{https://arxiv.org/abs/2402.18659}.

\bibitem[Khattab et~al.(2024)]{khattab2024dspy}
Omar Khattab et~al.
\newblock {DSPy}: Compiling declarative language model calls into
  self-improving pipelines.
\newblock In \emph{International Conference on Learning Representations
  (ICLR)}, 2024.
\newblock URL \url{https://openreview.net/pdf?id=sY5N0zY5Od}.

\bibitem[Koo and Li(2016)]{koo2016guideline}
Terry~K. Koo and Mae~Y. Li.
\newblock A guideline of selecting and reporting intraclass correlation
  coefficients for reliability research.
\newblock \emph{Journal of Chiropractic Medicine}, 15\penalty0 (2):\penalty0
  155--163, 2016.
\newblock \doi{10.1016/j.jcm.2016.02.012}.

\bibitem[Maleki and Zhao(2024)]{maleki2024pcg}
Mahdi~Farrokhi Maleki and Richard Zhao.
\newblock Procedural content generation in games: A survey with insights on
  emerging {LLM} integration.
\newblock \emph{arXiv preprint arXiv:2410.15644}, 2024.
\newblock URL \url{https://arxiv.org/abs/2410.15644}.

\bibitem[Mehenni and Zouaq(2024)]{mehenni2024ontology}
Gaya Mehenni and Amal Zouaq.
\newblock Ontology-constrained generation of domain-specific clinical
  summaries.
\newblock \emph{arXiv preprint arXiv:2411.15666}, 2024.
\newblock URL \url{https://arxiv.org/abs/2411.15666}.

\bibitem[Mei et~al.(2025)]{mei2025context}
Lingrui Mei et~al.
\newblock A survey of context engineering for large language models.
\newblock \emph{arXiv preprint arXiv:2507.13334}, 2025.
\newblock URL \url{https://arxiv.org/abs/2507.13334}.

\bibitem[Noy and McGuinness(2001)]{noy2001ontology}
Natalya~F. Noy and Deborah~L. McGuinness.
\newblock Ontology development 101: A guide to creating your first ontology.
\newblock Technical report, Stanford University, 2001.
\newblock URL
  \url{https://protege.stanford.edu/publications/ontology_development/ontology101.pdf}.

\bibitem[Shankar et~al.(2024)]{shankar2024validates}
Shreya Shankar et~al.
\newblock Who validates the validators? aligning {LLM}-assisted evaluation of
  {LLM} outputs with human preferences.
\newblock \emph{arXiv preprint arXiv:2404.12272}, 2024.
\newblock URL \url{https://arxiv.org/abs/2404.12272}.

\bibitem[Shrout and Fleiss(1979)]{shrout1979intraclass}
Patrick~E. Shrout and Joseph~L. Fleiss.
\newblock Intraclass correlations: Uses in assessing rater reliability.
\newblock \emph{Psychological Bulletin}, 86\penalty0 (2):\penalty0 420--428,
  1979.
\newblock \doi{10.1037/0033-2909.86.2.420}.

\bibitem[Toro et~al.(2024)]{toro2024dragon}
Sabrina Toro et~al.
\newblock {DRAGON-AI}: Dynamic retrieval augmented generation of ontologies
  using {AI}.
\newblock \emph{Journal of Biomedical Semantics}, 2024.
\newblock URL
  \url{https://jbiomedsem.biomedcentral.com/articles/10.1186/s13326-024-00320-3}.

\bibitem[Whitehead(1929)]{whitehead1929process}
Alfred~North Whitehead.
\newblock \emph{Process and Reality}.
\newblock Free Press, 1929.
\newblock Corrected edition, 1978.

\bibitem[Zheng et~al.(2024)]{zheng2024judging}
Lianmin Zheng et~al.
\newblock Judging {LLM}-as-a-judge with {MT-Bench} and chatbot arena.
\newblock \emph{Advances in Neural Information Processing Systems}, 36, 2024.
\newblock URL \url{https://arxiv.org/abs/2306.05685}.

\end{thebibliography}
}

% ============================================================
% APPENDICES
% ============================================================
\appendix

\section{Theme Prompts}
\label{app:prompts}

Table~\ref{tab:prompts} lists the ten standardized theme prompts used
in the ablation study.  Each prompt was run under all four experimental
conditions with three replications, yielding 120 total designs.

\begin{table*}[t]
\centering
\caption{Standardized theme prompts for the ablation study.}
\label{tab:prompts}
\small
\begin{tabularx}{\textwidth}{clX}
\toprule
\textbf{\#} & \textbf{Constraints} & \textbf{Theme} \\
\midrule
1  & 2--4p, competitive, medium, 45--60min     & Bioluminescent fungi competing in a cave ecosystem \\
2  & 2--4p, competitive, med-heavy, 60--90min   & Victorian-era detective solving mysteries through deduction \\
3  & 3--5p, competitive, med-heavy, 60--90min   & Space traders negotiating routes between planets \\
4  & 2--4p, competitive, light, 30--45min       & Medieval knights jousting in a tournament \\
5  & 2--4p, cooperative, medium, 45--60min      & Deep sea divers recovering treasure from a sunken ship \\
6  & 2--4p, competitive, medium, 45--60min      & Rival chefs competing in a cooking competition \\
7  & 2--5p, competitive, med-heavy, 90--120min  & Ancient civilizations building wonders of the world \\
8  & 2--4p, cooperative, medium, 30--45min      & Escape room where the walls are closing in \\
9  & 2--4p, competitive, medium, 45--60min      & Fairy tale creatures vying for control of an enchanted forest \\
10 & 2--4p, cooperative, heavy, 90--120min      & Time travelers preventing paradoxes across historical eras \\
\bottomrule
\end{tabularx}
\end{table*}

\section{Reference Games}
\label{app:games}

Table~\ref{tab:refgames} lists the 20 published board games used as
the reference set in Study~2.  Games were selected to span three
complexity tiers (light, medium, heavy) and represent widely recognized
titles with established design reputations.  Each game was
reverse-engineered into a complete GameGrammar ontology (25 required
fields plus 3 balance parameters) and evaluated through the identical
Design Coach pipeline used for generated designs.

\begin{table*}[t]
\centering
\caption{Reference games for the benchmark study (Study 2).}
\label{tab:refgames}
\small
\begin{tabularx}{\textwidth}{llXl}
\toprule
\textbf{\#} & \textbf{Game} & \textbf{Designer(s)} & \textbf{Tier} \\
\midrule
1  & Azul                     & Michael Kiesling                  & Light \\
2  & Codenames                & Vlaada Chv\'{a}til                & Light \\
3  & Kingdomino               & Bruno Cathala                     & Light \\
4  & Love Letter              & Seiji Kanai                       & Light \\
5  & Splendor                 & Marc Andr\'{e}                    & Light \\
\midrule
6  & 7 Wonders                & Antoine Bauza                     & Medium \\
7  & Catan                    & Klaus Teuber                      & Medium \\
8  & Concordia                & Mac Gerdts                        & Medium \\
9  & Dune: Imperium           & Paul Dennen                       & Medium \\
10 & Everdell                 & James A.\ Wilson                  & Medium \\
11 & Root                     & Cole Wehrle                       & Medium \\
12 & Spirit Island            & R.\ Eric Reuss                    & Medium \\
13 & Terraforming Mars        & Jacob Fryxelius                   & Medium \\
14 & Ticket to Ride           & Alan R.\ Moon                     & Medium \\
15 & Wingspan                 & Elizabeth Hargrave                 & Medium \\
\midrule
16 & Brass: Birmingham        & Gavan Brown \& Matt Tolman        & Heavy \\
17 & Gaia Project             & Jens Drögemüller \& Helge Ostertag & Heavy \\
18 & Gloomhaven               & Isaac Childres                    & Heavy \\
19 & Through the Ages         & Vl\'{a}cil Chv\'{a}til           & Heavy \\
20 & Twilight Imperium 4E     & Dane Beltrami \& Corey Konieczka  & Heavy \\
\bottomrule
\end{tabularx}
\end{table*}

\end{document}